\documentclass{article}
\usepackage{arxiv}
\usepackage{float}          % enforce figure's loc.
\usepackage[utf8]{inputenc} % allow utf-8 input
\usepackage[T1]{fontenc}    % use 8-bit T1 fonts
\usepackage{hyperref}       % hyperlinks
\usepackage{url}            % simple URL typesetting
\usepackage{booktabs}       % professional-quality tables
\usepackage{amsfonts}       % blackboard math symbols
\usepackage{nicefrac}       % compact symbols for 1/2, etc.
\usepackage{microtype}      % microtypography
\usepackage{lipsum}
\usepackage{graphicx}
\graphicspath{ {./images/} }

\title{Convolutional Neural Network based on transfer learning for Breast Cancer screening}

\author{
 Hussin Ragb \\
  Department of Electrical and Computer Engineering \\
  Christian Brothers University\\
   Memphis, TN 38104 \\
  \texttt{hragb@cbu.edu} \\
   \And
 Redha Ali \\
  Department of Electrical and Computer Engineering\\
  University of Dayton\\
  Dayton, Ohio 45469 \\
 \texttt{almahdir1@udayton.edu} \\
  \And
 Elforjani Jera \\
  Department of Electro Optics and Photonics\\
  University of Dayton\\
  Dayton, Ohio 45469 \\
  \texttt{ jerae1@udayton.edu} \\
   \And
 Nagi Buaossa \\
  Department of Electrical and Computer Engineering\\
  University of Dayton\\
  Dayton, Ohio 45469 \\
  \texttt{buaossan1@udayton.edu} \\
}

\begin{document}
\maketitle
\begin{abstract}
Breast cancer is the most common cancer in the world and the most prevalent cause of death among women worldwide. Nevertheless, it is also one of the most treatable malignancies if detected early. In this paper, a deep convolutional neural network-based algorithm is proposed to aid in accurately identifying  breast cancer from  ultrasonic images. In this algorithm, several neural networks are fused in a parallel architecture to perform the classification process and the voting criteria are applied in the final classification decision between the candidate object classes where the output of each neural network is representing a single vote. Several experiments were conducted on the breast ultrasound dataset consisting of 537 Benign, 360 malignant, and 133 normal images. These experiments show an optimistic result and a capability of the proposed model to outperform many state-of-the-art algorithms on several measures. Using k-fold cross-validation and a bagging classifier ensemble, we achieved an accuracy of 99.5\% and a sensitivity of 99.6\%.  
\end{abstract}

% keywords can be removed

\section{Introduction}
In today's world, breast cancer is considered one of the most common types of cancer and the incident rates keep increasing by 0.5\% per year. In 2020, an estimated 2.3 million women were diagnosed with breast cancer globally \cite{r1}. Breast cancer is caused by abnormal cells formed commonly in the cells of the lobules or the ducts. Cells typically have their cell growth which lets the old dying cells get replaced with healthy new cells. But, over time, these cells can mutate turning on oncogenes or turning off tumor suppressor genes. This causes the cell to uncontrollably divide producing more mutated cells and eventually forming a tumor. One of those tumors is a malignant tumor. Malignant tumors are considered cancerous and if left untreated, these malignant cells can spread throughout the body and can be life-threatening to the patient. Therefore, detecting breast cancer in the early stages can significantly increase the chances of survival. Meaning that it is vital to have proper screening methods to detect for initial symptoms of breast cancer.
Various imaging techniques for breast cancer detection include mammography, ultrasound, and thermography. Each technique has its pros and cons which justify its usage. Mammography has an average specificity of 75\% (25\% false-positive) and an average sensitivity of 80\% (20\% of cancers missed). Ultrasound has an average specificity of 66\% (34\% false-positive) and an average sensitivity of 83\% (17\% of cancers missed). Thermography has an average specificity of 90\% (10\% false-positive) and an average sensitivity of 90\% (10\% of cancers missed). The average specificity and average sensitivity can be improved using artificial intelligence. In the recent years, artificial intelligence has grown in popularity with the idea of using computers to perform tasks, simple or complex, that would usually be done by a human. In some cases, these artificial intelligence machines are capable of doing tasks that are not humanly possible or do a much better job by undergoing the process of machine learning. A convolutional neural network (CNN) is a type of artificial neural network which is primarily used for image recognition and processing because of its ability to recognize patterns in images very well. Due to its ability to recognize patterns in images very well, it has gotten popular in the health department such as in the detection of breast cancer. Other studies \cite{r1, r2,r3,r4,r5,r6,r7,r8,r9,r10,r11,r12,r13,r14,r31, r33} have proposed artificial intelligence and CNN as an imaging technique because it has higher average specificity and sensitivity while also being more efficient than other imaging techniques. In this paper, we fine-tune nine neural networks pre-trained on the ImageNet dataset with the fully connected layers replaced; these networks are VGG-16 \cite{r16} , Xception \cite{r15}, ResNet-18, ResNet-50 \cite{r17}, DenseNet201 \cite{r18}, GoogleNet \cite{r19}, AlexNet \cite{r20}, MobileNet \cite{r21}, and Darknet19 \cite{r22}. These networks are fine-tuned and retrained on the ultrasound breast cancer images and the performance parameters such as accuracy, sensitivity, specificity, precision, and area under the curve (AUC) are computed for  each of these neural networks. A combination of four fine-tuned pre-trained networks that have the highest performance are fused in a parallel architecture to perform the classification process and the voting criteria are applied in the final classification decision between the candidate object classes where the output of each neural network is representing a single vote.

% \subsection{ text is here}
\section{Material and method}
\label{sec:headings}
\subsection{Dataset}

\begin{figure}[tbh]
  \centering
    \includegraphics[scale=0.6]{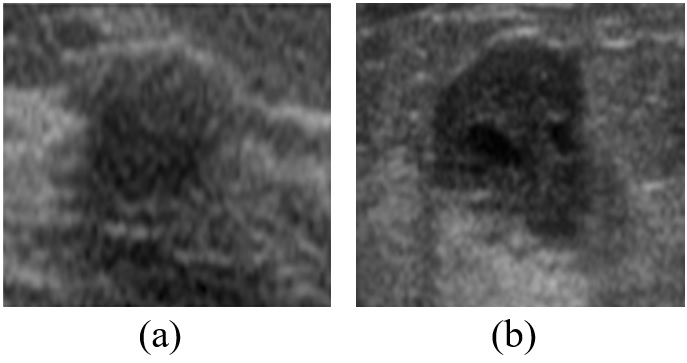}
    \caption{Samples of the ultrasonic images from dataset 1: 
    (a) Benign breast cancer.  (b) Malignant breast cancer.}
    \label{fig:fig1}
\end{figure}

\begin{figure}[tbh]
  \centering
    \includegraphics[scale=0.5]{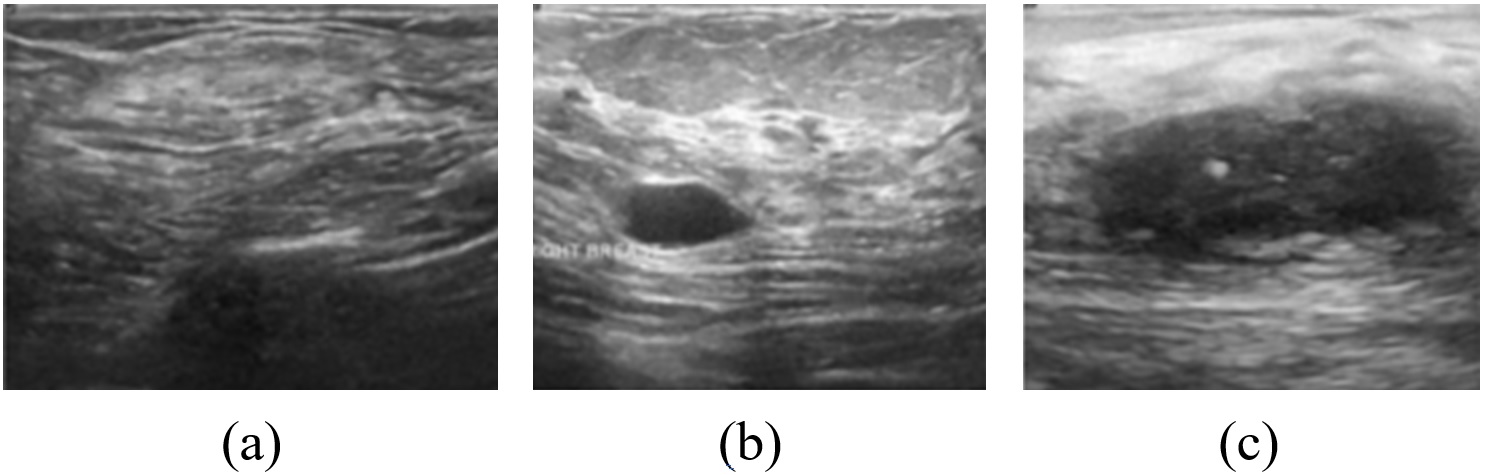}
    \caption{Samples of the ultrasonic images from the dataset 2: (a) Normal (b) Benign cancer.  (c) Malignant cancer.}
    \label{fig:fig2}
\end{figure}

In this research, two breast ultrasound image datasets    \cite{r23,r24} were considered. The dataset in \cite{r23} contains 250 images in which there are two categories: malignant and benign cases. The images in this dataset have different sizes with gray and RGB colors. The minimum and the maximum size of the images are $57 \times 75$ and $61 \times 199$ pixels, respectively. The dataset in \cite{r24} contains 780 images, in which there are three categories: malignant, benign, and normal cases. These images are collected from 600 women in 2018, and the age range of the women is between 25 and 75 years. The average size of the images in this dataset is $500 \times  500$ pixels. The images in \cite{r23,r24} are combined to increase the size of the training and testing dataset. That’s to avoid the overfitting and biases in the training process and to consider the three classes (benign, malignant, and normal) at the output of the neural network. Samples of the ultrasonic images from  dataset 1 \cite{r23} and dataset 2 \cite{r24} are illustrated in Figure \ref{fig:fig1} and Figure \ref{fig:fig2} respectively. The class distribution of the images in the two datasets is shown in Table \ref{table:table1}.

%Table~\ref{tab:table1} 

% Figure-1 start here

% Figure-2 start here

% Table-1 start here
\begin{table}[htp]
\caption{Datasets and distribution of classes   }
\centering
\begin{tabular}{|c|c|c|c|c|c|c|c|}
\hline
Dataset & Normal & Benign & Malignant & Total  \\ \hline
Dataset 1 &  -  &  100 &  150 &  250 \\
Dataset 2 &  133  &  437 &  210  &  780 \\
Total &  133 &  537 &  360 &  1030 \\\hline
\end{tabular}
\label{table:table1}
\end{table}

\subsection{Data pre-processing}
To ensure the robustness of  each model in this paper and to prevent overfitting on the datasets \cite{r23,r24}, we introduce data augmentation in the form of the random x-axis and y-axis reflections with 50\% probability, 360-degree random rotation, random x-axis, and y-axis translation between the range of -30 to 30 pixels, and an x-axis and y-axis scale range between 0.9 and 1.1 magnitude \cite{r25}.

\subsection{Network models}
Each model in Table \ref{table:table2} \cite{r29,r30}is trained and evaluated for  use in ensemble combinations. Four models that were pre-trained on the ImageNet Dataset, are fine-tuned with the Breast Cancer dataset \cite{r23,r24} with final layers being replaced. The four models are VGG16, GoogleNet, ResNet-18, and ResNet-50. Experiments are then categorized to determine an individual’s robustness with the ensembled network combinations.

% Table-2 start here
\begin{table}[htp]
\caption{Datasets and distribution of classes   }
\centering
\begin{tabular}{|c|c|c|c|c|c|c|c|}
\hline
Model & Input size & Parameters(Millions) & Depth \\ \hline
MobileNet    & $224 \times 224$   &  3.5 &    53  \\
GoogleNet    & $224 \times 224$   &  7   &    22   \\
ResNet18     & $224 \times 224$   & 11.7 &    18  \\
DarkNet19    & $256 \times 256$   & 21   &    19   \\
AlexNet      & $227 \times 227$   & 61   &    8    \\
Xception     & $229 \times 229$   & 22.9 &    71    \\
ResNet50     & $224 \times 224$   & 25.6 &    50    \\
DensNet201   & $224 \times 224$   & 20   &    201  \\
VGG16        & $224 \times 224$   & 138  &    16    \\\hline
\end{tabular}
\label{table:table2}
\end{table}

\subsection{Ensemble Learning}
\begin{figure}[tbh]
  \centering
    \includegraphics[scale=0.6]{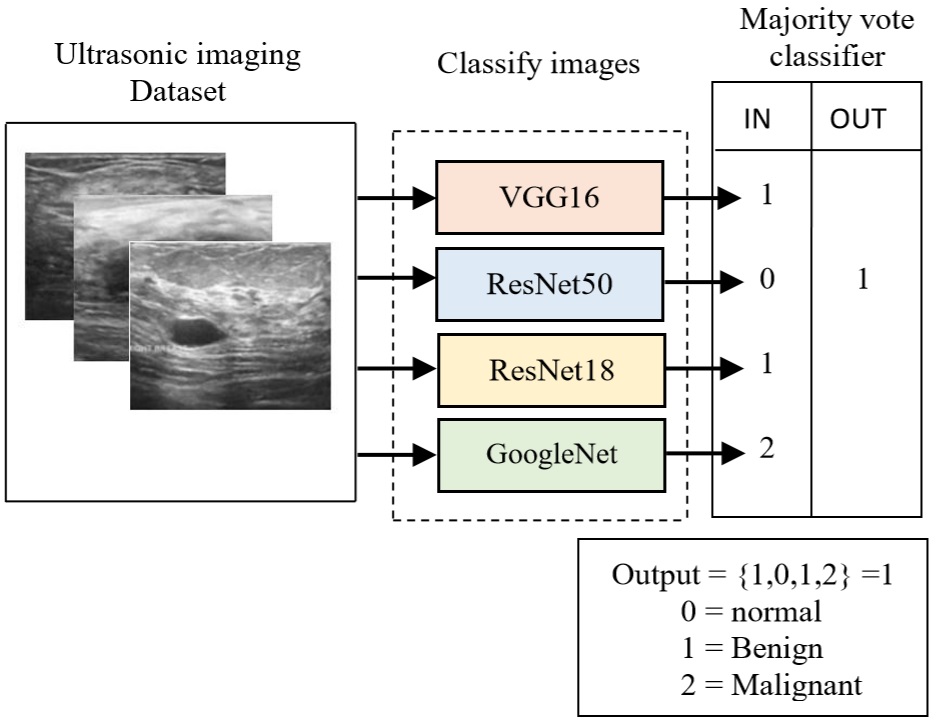}
    \caption{Bagging Approach to Ensemble Learning using a Majority Vote Classifier}
    \label{fig:fig3}
\end{figure}

\begin{figure}[tbh]
  \centering
    \includegraphics[scale=0.6]{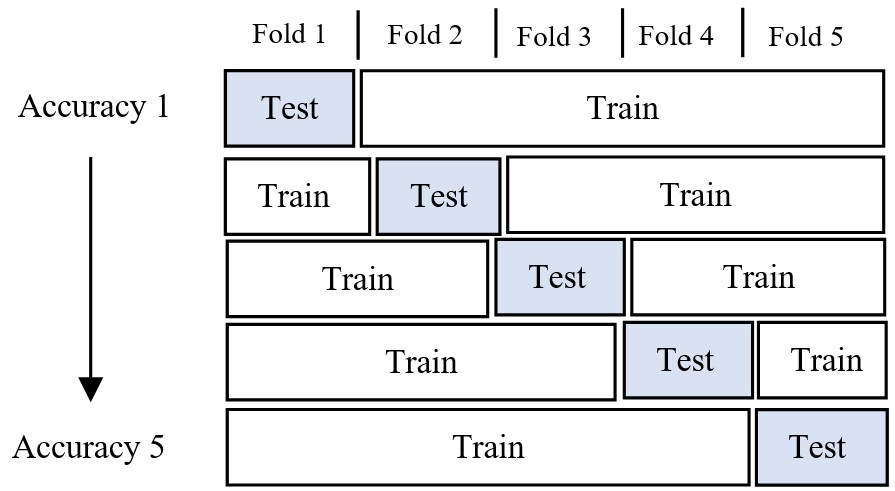}
    \caption{Illustration of K-fold Cross-validation (k = 1-5)}
    \label{fig:fig4}
\end{figure}

Classifiers whose predictions have combined are known as Ensemble. Research has proven Ensemble to have an increase in accuracy when compared to other individual classifiers for various imaging applications \cite{r26,r27,r32,r34,r37}. Bagging Classifiers and Boosting Classifiers are the two most popular methods of ensemble learning. Bagging classifiers train each classifier on varying subsets of the target dataset. It is most effective in unstable neural networks in which small changes in the training set cause significant changes in test set predictions \cite{r28}. Moreover, each network in the ensemble may have an innate propensity to predict different subsets of the dataset more accurately. As a result, Bagging Classifiers typically perform better than any other individual classifier in the ensemble \cite{r28}.\\
Boosting Classifiers is a group of methods that involves altering the input dataset to the training phases of proceeding networks in an ensemble based on the performance of preceding networks. This is the practice of tailoring the input datasets of networks such that an optimal dataset representation is achieved that is non-reliant on input variance across the ensemble classifier. This ensemble method can vastly outperform bagging classifiers, but it has the potential to perform worse than an individual network in the classifier. Boosting classifier performance is heavily dependent on features of the target dataset \cite{r28}. 
 For this experiment, the bagging approach is adopted to maximize performance. Each network is trained using k-fold cross-validation. The mean accuracy is measured across each fold of the trained models as seen in Figure \ref{fig:fig4}. The output classification of the three selected networks is then grouped into a majority vote classifier ensemble neural network. A majority vote classifier takes the mode of the binary classifications as seen in Figure \ref{fig:fig3} \cite{r29, r30}.

% Figure-3 start here

\subsection{Experimental setup}
MATLAB Programming Language was used to train the transferring learning models. VGG16, GoogleNet, ResNet18, and ResNet50 are all CNN Models, which were pretrained on a subset of the ImageNet dataset which was used in the ImageNet Large-Scale Visual Recognition Challenge (ILSVRC), which consists of thousands of class labels and over a million images. For each network, the last three layers – fully connected, SoftMax, and classification layer are replaced with the weights randomly initialized. Each network was trained for 15 epochs with a batch size and learning rate of 8 and 0.00005, respectively. Both, stochastic `gradient descent with momentum (SGDM), optimizer and adaptive moment estimation (Adam) optimizers are used in the training process.  The learning rate that is used for models’ training is constant, shuffle is performed each epoch. As a cross-validation method, k-fold is chosen as a strategy to combat the limited data samples. The dataset split was 20\% test set and 80\% training set in each fold. The K-fold cross-validation is used to ensure that each observation from the raw dataset can appear in both the training and testing set. Results were obtained using 5 different k values (1-5). Then each network is combined into a bagging ensemble neural network. This bagging ensemble uses a majority vote classifier which takes the mode of each output of the combined neural networks as seen in Figure \ref{fig:fig3}. It is expected that a bagging ensemble will outperform any individual network model.

% Figure-4 start here

\subsection{Performance metrics}
The measurement parameters TP, FP, TN, and FN represent True Positive, False Positive, True Negative, and False Negative percentages, respectively. 
Diagnostic testing necessitates a low False Negative (FN) rate or high sensitivity often at the risk of increasing the False Positive (FP) rate. 
The effectiveness of each neural network is evaluated with various standard metrics of performance. 
These metrics are the accuracy, precision, specificity, recall, sensitivity, and F1- score, False Positive Rate (FPR), and the Area Under Curve (AUC) \cite{r29,r30}. These metrics are defined as follows:\\

% ((((  Equations start here  ))))

\begin{equation}
  Accuracy =  { \frac{(TP+TN)}{(TP+TN+FP+FN)}}
\label{Accuracy_eq}
\end{equation}

\begin{equation}
  Precision =  { \frac{TP}{(TP+FP)}}
\label{ Precision_eq}
\end{equation}

\begin{equation}
   Specificity   =  { \frac{TN}{(TN+FP)}}
\label{Specificity_eq}
\end{equation}

\begin{equation}
   Sensitivity    =  { \frac{TP}{(TP+FN)}}
\label{Sensitivity_eq}
\end{equation}

\begin{equation}
   F1_{score}     =  { \frac{2 (precision) (recall)}{(precision + recall)}}
\label{F1_eq}
\end{equation}

\begin{equation}
   False Positive Rate (FPR)    =  { \frac{FP}{(FP+TN)}}
\label{FPR_eq}
\end{equation}

\begin{equation}
   Negative Predictive Value     =  { {TN}{(TN+FN)}}
\label{NPR_eq}
\end{equation}

\begin{figure}[tbh]
  \centering
    \includegraphics[scale=0.6]{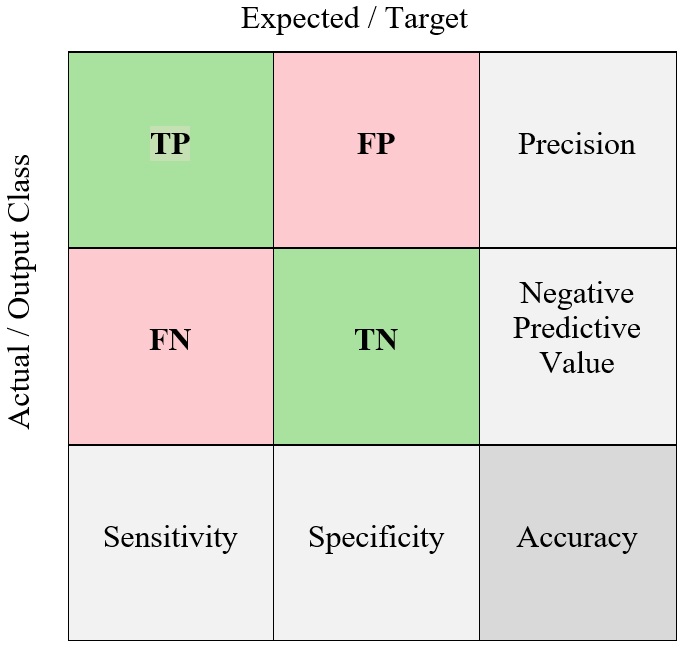}
    \caption{Confusion matrix}
    \label{fig:fig5}
\end{figure}

These metrics of performance are represented in the confusion matrix that is shown in Figure \ref{fig:fig5}. 
For medical diagnostic purposes, it is imperative that all positive cases are correctly classified. Therefore, a lower decision threshold is preferred. At each decision threshold, the sensitivity or true positive rate (TPR) as found in Equation (4) is plotted in the y-axis against the ‘average False Positive per image’ or false positive rate (FPR) as found in Equation (6). As the threshold is lowered, both the TPR and FPR increase or remain constant. The area under the ROC curve (AUC) aggregates the performance of the neural network on all decision thresholds. This represents the average performance of the neural network. The closer the AUC is to 1.0, the better the network is at distinguishing between positive and negative classes \cite{r29,r30,r35,r36}.

% Figure-5 start here

\section{Results and discussion}
\begin{figure}[H]
  \centering
    \includegraphics[scale=0.7]{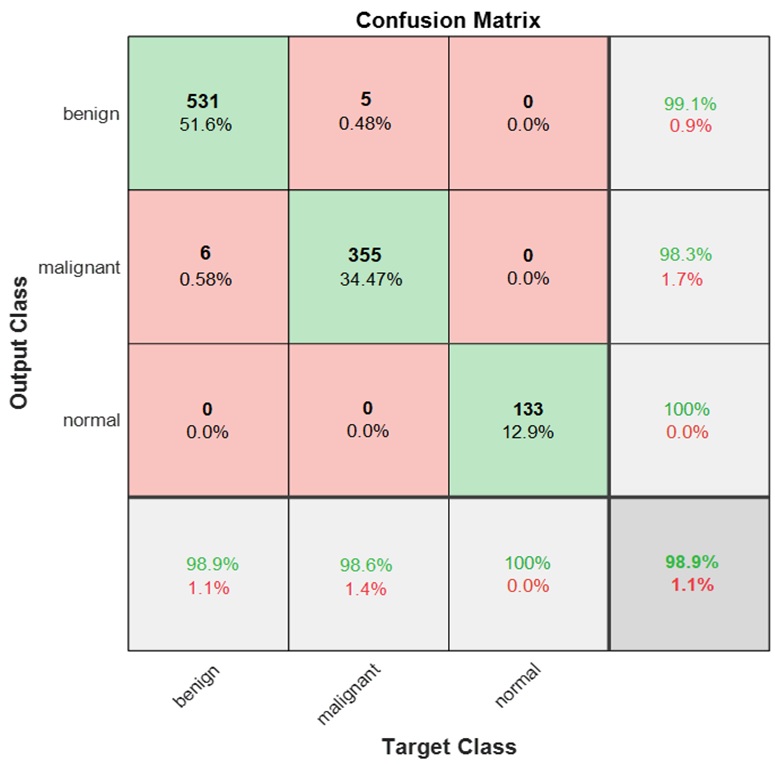}
    \caption{Confusion matrix of the proposed model using SGDM optimizer}
    \label{fig:fig6}
\end{figure}

\begin{figure}[H]
  \centering
    \includegraphics[scale=0.7]{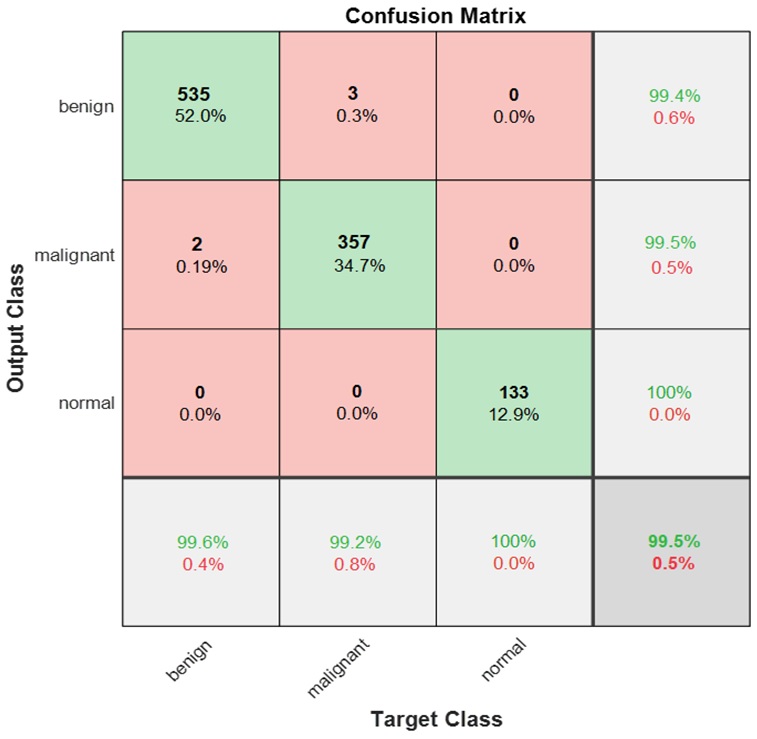}
    \caption{Confusion matrix of the proposed model using (Adam optimizer)}
    \label{fig:fig7}
\end{figure}

In this research work, AlexNet, GoogleNet, DenseNet201, VGG16, DarkNet19, MobileNet, Xception, ResNet18, and ResNet50 fine-tuned neural networks have been trained with the normal and breast cancer images  from the datasets \cite{r23,r24}. The metrics of the performance, accuracy, sensitivity, specificity, precision, and area under the curve (AUC) were computed for each of these fine-tuned pre-trained neural networks as well as the proposed model as shown in Table 3 and Table 4 using SGDM optimizer and Adam optimizer respectively. Among the pre-trained neural network, ResNet50, GoogleNet, ResNet18, and VGG16 demonstrated the highest performance with accuracy scores  93.1\%, 90.5\%, 90.4\%, and 90.2\% respectively using Adam optimizer. These neural networks were selected and used in the proposed ensemble model. The results illustrated in Table \ref{table:table3} and Table \ref{table:table4} showed the outperform of the proposed model over all the single fine-tuned trained neural networks. The accuracy of the ensemble model was 98.9\% for the SGDM optimizer and 99.5\% for the Adam optimizer. The confusion matrix of the proposed model using the SGDM, and Adam optimizers are illustrated in Figure \ref{fig:fig6} and Figure \ref{fig:fig7} respectively.

% Table-3 start here
\begin{table}[htp]
\caption{Evaluation results of pre-trained neural networks and the proposed model using (SGDM optimizer)  }
\centering
\begin{tabular}{|l|c|c|c|c|c|c|c|}
\hline
Model & Sensitivity & Specificity & Accuracy & AUC \\ \hline
1.  AlexNet    &  90.7\%   &  88.4\%   &   89.6\%  & 0.965 \\
2.  GoogleNet   & 91.2\%   &  86.5\%   &   88.9\%  & 0.954 \\
3.  DensNet201  & 90.3\%   &  88.7\%   &   89.1\%  & 0.948 \\
4.  VGG16       & 91.9\%   &  87.6\%   &   89.8\%  & 0.966  \\
5.  DarkNet19   & 88.3\%   &  85.9\%   &   87.2\%  & 0.946  \\
6.  MobileNet   & 85.1\%   &  86.4\%   &   85.7\%  & 0.939 \\
7.  Xception    & 83.7\%   &  82.2\%   &   83.1\%  & 0.919  \\
8.  ResNet18    & 90.1\%   &  85.1\%   &   87.8\%  & 0.944  \\
9.  ResNet50    & 83.5\%   &  88.5\%   &   87.5\%  & 0.954 \\
Our model, [fusing(2,4,8,9)] & 98.9\%  & 98.6\% & 98.9\% & 0.989   \\\hline
\end{tabular}
\label{table:table3}
\end{table}

% Figure-6 start here

% Table-4 start here
\begin{table}[htp]
\caption{Evaluation results of pre-trained neural networks and the proposed model using (Adam optimizer)}
\centering
\begin{tabular}{|l|c|c|c|c|c|c|c|}
\hline
Model & Sensitivity & Specificity & Accuracy & AUC \\ \hline
1.  AlexNet     & 85.8\%   &  83.9\%   &   84.9\%  & 0.939 \\
2.  GoogleNet   & 90.5\%   &  86.6\%   &   90.5\%  & 0.962 \\
3.  DensNet201  & 89.2\%   &  87.6\%   &   88.7\%  & 0.951 \\
4.  VGG16       & 92.9\%   &  86.7\%   &   90.2\%  & 0.973  \\
5.  DarkNet19   & 88.4\%   &  85.9\%   &   87.2\%  & 0.946  \\
6.  MobileNet   & 88.9\%   &  90.9\%   &   89.9\%  & 0.957 \\
7.  Xception    & 89.9\%   &  91.2\%   &   82.9\%  & 0.952  \\
8.  ResNet18    & 91.2\%   &  89.6\%   &   90.4\%  & 0.969  \\
9.  ResNet50    & 94.1\%   &  92.1\%   &   93.1\%  & 0.976 \\
Our model, [fusing(2,4,8,9)] & 99.6\%  & 99.2\% & 99.5\% & 0.992   \\\hline
\end{tabular}
\label{table:table4}
\end{table}

% Figure-7 start here

\section{Conclusion}
In this paper, nine fine-tuned pre-trained CNN neural networks were implemented to observe their effectiveness on the classification of breast cancer from ultrasound images that combined from two datasets. ResNet50, GoogleNet, ResNet18, and VGG16 models have registered the best performance for detecting  breast cancer. In our proposed model, these neural networks were fused, and the voting criteria were applied in the final classification decision between the candidate object classes where the output of each neural network was representing a single vote. The proposed model achieved 99.5\% accuracy, 99.6\% sensitivity and, 0.992 AUC value.

\bibliographystyle{unsrt}  
\bibliography{references}  %%% Remove comment to use the external .bib file (using bibtex).

\begin{thebibliography}{10}

\bibitem{r1}
J~Arevalo.
\newblock Gonz alez fa, ramos-poll an r, oliveira jl, guevara lopez ma.
  representation learning for mammography mass lesion classification with
  convolutional neural networks.
\newblock {\em Comput Methods Programs Biomed}, 127:248--57, 2016.

\bibitem{r2}
Benjamin~Q Huynh, Hui Li, and Maryellen~L Giger.
\newblock Digital mammographic tumor classification using transfer learning
  from deep convolutional neural networks.
\newblock {\em Journal of Medical Imaging}, 3(3):034501, 2016.

\bibitem{r3}
Yuan-Pin Lin and Tzyy-Ping Jung.
\newblock Improving eeg-based emotion classification using conditional transfer
  learning.
\newblock {\em Frontiers in human neuroscience}, 11:334, 2017.

\bibitem{r4}
Christian Szegedy, Wei Liu, Yangqing Jia, Pierre Sermanet, Scott Reed, Dragomir
  Anguelov, Dumitru Erhan, Vincent Vanhoucke, and Andrew Rabinovich.
\newblock Going deeper with convolutions.
\newblock In {\em Proceedings of the IEEE conference on computer vision and
  pattern recognition}, pages 1--9, 2015.

\bibitem{r5}
Neil~J Vickers.
\newblock Animal communication: when i’m calling you, will you answer too?
\newblock {\em Current biology}, 27(14):R713--R715, 2017.

\bibitem{r6}
Hiroki Tanaka, Shih-Wei Chiu, Takanori Watanabe, Setsuko Kaoku, and Takuhiro
  Yamaguchi.
\newblock Computer-aided diagnosis system for breast ultrasound images using
  deep learning.
\newblock {\em Physics in Medicine \& Biology}, 64(23):235013, 2019.

\bibitem{r7}
Neil~J Vickers.
\newblock Animal communication: when i’m calling you, will you answer too?
\newblock {\em Current biology}, 27(14):R713--R715, 2017.

\bibitem{r8}
Walid Al-Dhabyani, Mohammed Gomaa, Hussien Khaled, and Fahmy Aly.
\newblock Deep learning approaches for data augmentation and classification of
  breast masses using ultrasound images.
\newblock {\em Int. J. Adv. Comput. Sci. Appl}, 10(5):1--11, 2019.

\bibitem{r9}
Christian Szegedy, Vincent Vanhoucke, Sergey Ioffe, Jon Shlens, and Zbigniew
  Wojna.
\newblock Rethinking the inception architecture for computer vision.
\newblock In {\em Proceedings of the IEEE conference on computer vision and
  pattern recognition}, pages 2818--2826, 2016.

\bibitem{r10}
Barret Zoph, Vijay Vasudevan, Jonathon Shlens, and Quoc~V Le.
\newblock Learning transferable architectures for scalable image recognition.
\newblock In {\em Proceedings of the IEEE conference on computer vision and
  pattern recognition}, pages 8697--8710, 2018.

\bibitem{r11}
Ting Xiao, Lei Liu, Kai Li, Wenjian Qin, Shaode Yu, and Zhicheng Li.
\newblock Comparison of transferred deep neural networks in ultrasonic breast
  masses discrimination.
\newblock {\em BioMed research international}, 2018, 2018.

\bibitem{r12}
Heqing Zhang, Lin Han, Ke~Chen, Yulan Peng, and Jiangli Lin.
\newblock Diagnostic efficiency of the breast ultrasound computer-aided
  prediction model based on convolutional neural network in breast cancer.
\newblock {\em Journal of Digital Imaging}, 33:1218--1223, 2020.

\bibitem{r13}
Abdullah-Al Nahid and Yinan Kong.
\newblock Histopathological breast-image classification using local and
  frequency domains by convolutional neural network.
\newblock {\em Information}, 9(1):19, 2018.

\bibitem{r14}
Abdullah-Al Nahid, Mohamad~Ali Mehrabi, and Yinan Kong.
\newblock Histopathological breast cancer image classification by deep neural
  network techniques guided by local clustering.
\newblock {\em BioMed research international}, 2018, 2018.

\bibitem{r31}
Mehedi Masud, Amr E~Eldin Rashed, and M~Shamim Hossain.
\newblock Convolutional neural network-based models for diagnosis of breast
  cancer.
\newblock {\em Neural Computing and Applications}, pages 1--12, 2020.

\bibitem{r33}
Barath~Narayanan Narayanan, Vignesh Krishnaraja, and Redha Ali.
\newblock Convolutional neural network for classification of histopathology
  images for breast cancer detection.
\newblock In {\em 2019 IEEE National Aerospace and Electronics Conference
  (NAECON)}, pages 291--295. IEEE, 2019.

\bibitem{r16}
Karen Simonyan and Andrew Zisserman.
\newblock Very deep convolutional networks for large-scale image recognition.
\newblock {\em arXiv preprint arXiv:1409.1556}, 2014.

\bibitem{r15}
Fran{\c{c}}ois Chollet.
\newblock Xception: Deep learning with depthwise separable convolutions.
\newblock In {\em Proceedings of the IEEE conference on computer vision and
  pattern recognition}, pages 1251--1258, 2017.

\bibitem{r17}
Kaiming He, Xiangyu Zhang, Shaoqing Ren, and Jian Sun.
\newblock Deep residual learning for image recognition [c].
\newblock In {\em Proceedings of the IEEE conference on computer vision and
  pattern recognition}, volume 2016, pages 770--778, 2016.

\bibitem{r18}
Gao Huang, Zhuang Liu, Laurens Van Der~Maaten, and Kilian~Q Weinberger.
\newblock Densely connected convolutional networks.
\newblock In {\em Proceedings of the IEEE conference on computer vision and
  pattern recognition}, pages 4700--4708, 2017.

\bibitem{r19}
Christian Szegedy, Wei Liu, Yangqing Jia, Pierre Sermanet, Scott Reed, Dragomir
  Anguelov, Dumitru Erhan, Vincent Vanhoucke, and Andrew Rabinovich.
\newblock Going deeper with convolutions.
\newblock In {\em Proceedings of the IEEE conference on computer vision and
  pattern recognition}, pages 1--9, 2015.

\bibitem{r20}
Alex Krizhevsky, Ilya Sutskever, and Geoffrey~E Hinton.
\newblock Imagenet classification with deep convolutional neural networks.
\newblock {\em Advances in neural information processing systems},
  25:1097--1105, 2012.

\bibitem{r21}
Andrew~G Howard, Menglong Zhu, Bo~Chen, Dmitry Kalenichenko, Weijun Wang,
  Tobias Weyand, Marco Andreetto, and Hartwig Adam.
\newblock Mobilenets: Efficient convolutional neural networks for mobile vision
  applications.
\newblock {\em arXiv preprint arXiv:1704.04861}, 2017.

\bibitem{r22}
Joseph Redmon and Ali Farhadi.
\newblock Yolo9000: better, faster, stronger.
\newblock In {\em Proceedings of the IEEE conference on computer vision and
  pattern recognition}, pages 7263--7271, 2017.

\bibitem{r23}
Paulo~Sergio Rodrigues.
\newblock Breast ultrasound image.
\newblock {\em Mendeley Data}, 1, 2017.

\bibitem{r24}
Neil~J Vickers.
\newblock Animal communication: when i’m calling you, will you answer too?
\newblock {\em Current biology}, 27(14):R713--R715, 2017.

\bibitem{r25}
Jason Yosinski, Jeff Clune, Yoshua Bengio, and Hod Lipson.
\newblock How transferable are features in deep neural networks?
\newblock {\em arXiv preprint arXiv:1411.1792}, 2014.

\bibitem{r29}
Hussin~K Ragb, Ian~T Dover, and Redha Ali.
\newblock Fused deep convolutional neural network for precision diagnosis of
  covid-19 using chest x-ray images.
\newblock {\em arXiv preprint arXiv:2009.08831}, 2020.

\bibitem{r30}
Hussin~K Ragb, Ian~T Dover, and Redha Ali.
\newblock Deep convolutional neural network ensemble for improved malaria
  parasite detection.
\newblock In {\em 2020 IEEE Applied Imagery Pattern Recognition Workshop
  (AIPR)}, pages 1--10. IEEE, 2020.

\bibitem{r26}
Redha Ali and Hussin~K Ragb.
\newblock Fused deep convolutional neural networks based on voting approach for
  efficient object classification.
\newblock In {\em 2019 IEEE National Aerospace and Electronics Conference
  (NAECON)}, pages 335--339. IEEE, 2019.

\bibitem{r27}
Redha Ali, Russell~C Hardie, Barath~Narayanan Narayanan, and Supun De~Silva.
\newblock Deep learning ensemble methods for skin lesion analysis towards
  melanoma detection.
\newblock In {\em 2019 IEEE National Aerospace and Electronics Conference
  (NAECON)}, pages 311--316. IEEE, 2019.

\bibitem{r32}
Redha Ali, Russell~C Hardie, and Hussin~K Ragb.
\newblock Ensemble lung segmentation system using deep neural networks.
\newblock In {\em 2020 IEEE Applied Imagery Pattern Recognition Workshop
  (AIPR)}, pages 1--5. IEEE, 2020.

\bibitem{r34}
Redha Ali, Russell Hardie, and Almabrok Essa.
\newblock A leaf recognition approach to plant classification using machine
  learning.
\newblock In {\em NAECON 2018-IEEE National Aerospace and Electronics
  Conference}, pages 431--434. IEEE, 2018.

\bibitem{r37}
Hussin Ragb and Vijayan Asari.
\newblock Multi-hypothesis approach for efficient human detection.
\newblock {\em Journal of Imaging Science and Technology}, 63(2):20503--1,
  2019.

\bibitem{r28}
David Opitz and Richard Maclin.
\newblock Popular ensemble methods: An empirical study.
\newblock {\em Journal of artificial intelligence research}, 11:169--198, 1999.

\bibitem{r35}
Hussin~K Ragb and Vijayan~K Asari.
\newblock Histogram of oriented phase (hop): a new descriptor based on phase
  congruency.
\newblock In {\em Mobile Multimedia/Image Processing, Security, and
  Applications 2016}, volume 9869, page 98690V. International Society for
  Optics and Photonics, 2016.

\bibitem{r36}
Hussin~K Ragb and Vijayan~K Asari.
\newblock Histogram of oriented phase and gradient (hopg) descriptor for
  improved pedestrian detection.
\newblock {\em Electronic Imaging}, 2016(3):1--6, 2016.

\end{thebibliography}


\end{document}